\documentclass{article}

\usepackage[utf8]{inputenc}
\usepackage[T1]{fontenc}
\usepackage{hyperref}
\usepackage{url}
\usepackage{booktabs}
\usepackage{multirow}
\usepackage{amsfonts}
\usepackage{amsmath}
\usepackage{amssymb}
\usepackage{nicefrac}
\usepackage{microtype}
\usepackage[table]{xcolor}
\usepackage{graphicx}
\usepackage{subfig}
\usepackage{natbib}

\definecolor{directbg}{RGB}{248,244,255}
\definecolor{implicitbg}{RGB}{242,248,242}
\definecolor{meanbg}{RGB}{245,245,245}
\definecolor{swagbg}{RGB}{248,244,238}
\definecolor{highblue}{RGB}{226,236,248}
\definecolor{lowred}{RGB}{248,232,232}
\definecolor{midgray}{RGB}{243,243,243}
\definecolor{dashgray}{RGB}{150,150,150}

\title{Unstable Rankings in Bayesian Deep Learning Evaluation}

\author{
Qishi Zhan\thanks{Corresponding author.}\\
Department of Mathematical and Statistical Sciences\\
Marquette University, USA\\
\texttt{qishi.zhan@marquette.edu}
\and
Minxuan Hu\\
Cornell University, USA
\and
Guansu Wang\\
The University of Melbourne, Australia
\and
Jiaxin Liu\\
University of Illinois Urbana-Champaign, USA
\and
Liang He\\
Tongji University, China
}

\begin{document}

\maketitle

\begin{abstract}
Standard evaluations of Bayesian deep learning methods assume that
metric estimates are reliable, but we show this assumption fails
under data scarcity. Method rankings are not only unreliable at
small $n$, but also dataset-dependent in ways that point estimates
cannot reveal: the same method comparison yields
$P(\mathrm{MCD} \prec \mathrm{Ensemble}) = 1.000$ at $n = 50$
on one dataset and remains below $0.95$ even at $n = 500$ on
another. Across the datasets we consider, no universal sample size
threshold exists, which is precisely why dataset-specific posterior
inference is necessary. To address this, we use a Bayesian
hierarchical model with method-specific variances to treat
evaluation metrics as random variables across data realizations, and
we use a predictive Minimum Detectable Difference curve to assess
whether an observed gap would be detectable at a given training
size. Across six Bayesian deep learning methods and five regression
datasets, our results show that uncertainty-aware evaluation is
necessary in low-data settings, because current evidence for method
superiority and predictive detectability at the same training size
can diverge substantially. Our framework provides practitioners with
principled tools to determine whether their evaluation data is
sufficient before drawing conclusions about method superiority.
\end{abstract}

\section{Introduction}

Evaluating Bayesian deep learning methods requires more than accurate predictions.
It requires reliable uncertainty estimates.
Recent large-scale studies have benchmarked modern methods including Monte Carlo
Dropout, Deep Ensembles, Stochastic Weight Averaging Gaussian, and variational
inference across diverse architectures and tasks
\citep{seligmann2023beyond, ovadia2019trust, kirchhof2023url, mucsanyi2024benchmarking},
establishing performance hierarchies that the community now widely accepts.
These studies share a common assumption: training sets contain thousands of
examples.
The smallest split in \citet{seligmann2023beyond} comprises several thousand
samples.

This assumption fails silently in drug discovery, rare disease diagnosis,
medical imaging, and environmental monitoring, where labeled data is scarce
by nature and Bayesian or uncertainty-aware models are often deployed in
specialized settings
\citep{svensson2025enhancing,bargagna2023bayesian,ferchichi2025environmental,yu2025comparative}.
In these settings, a question that has received no systematic attention arises.
When training data is limited, are the metrics used to rank Bayesian deep
learning methods themselves reliable?

We answer this question empirically, and the answer is often no. On the
Concrete dataset, the posterior probability that Monte Carlo Dropout
outperforms Deep Ensembles on the Continuous Ranked Probability Score is
0.450 at $n = 200$, statistically indistinguishable from chance, and remains
below 0.95 even at $n = 500$. Reliable discrimination between these two
methods requires substantially more training samples than practitioners
typically assume, with thresholds that vary across datasets and method pairs.
The perceived advantage of leading methods over weaker baselines frequently
falls within the noise of metric estimation, making point-estimate
comparisons scientifically unreliable.

The source of this problem is not the methods themselves but the metrics used
to evaluate them.
Standard scoring rules such as the Continuous Ranked Probability Score and
Negative Log-Likelihood are estimated from a single draw of $n$ training
samples.
When $n$ is small, the variance of these estimators is large enough that
method rankings reflect sampling noise as much as true performance
differences.
Existing work on statistical evaluation in machine learning has examined
variance due to training randomness
\citep{bouthillier2021accounting, dehghani2021benchmark, raji2021everything, reuel2024betterbench, longjohn2024repositories}
and uncertainty in aggregate metrics for foundation model benchmarks
\citep{longjohn2025statistical}, but neither addresses the regime where
training data itself is the scarce resource.

To address this gap, we propose a Bayesian hierarchical model that treats
evaluation metrics as random variables across data realizations, enabling
posterior inference over pairwise method comparisons.
We further introduce the Minimum Detectable Difference curve, which quantifies
the smallest true performance gap that can be reliably detected given $n$
training samples, providing a practical sample size diagnostic for
practitioners.

Through a systematic study of six representative Bayesian deep learning
methods across five training set sizes on one synthetic and four real-world
regression datasets, we establish three findings.
First, standard UQ metrics decrease in variance with $n$, but at rates that
vary substantially across methods, because model training randomness
compounds sampling variance in ways that standard analysis ignores.
Second, method rankings are not stable across datasets: conclusions that hold
on one dataset reverse on another, and point-estimate comparisons cannot
detect these reversals.
Third, coverage-based metrics such as the Prediction Interval Coverage
Probability remain unreliable below $n = 100$, and among proper scoring
rules, CRPS provides substantially more stable small-sample estimates than
Negative Log-Likelihood.

Taken together, these results show that the performance hierarchies
established by large-scale benchmark studies cannot be assumed to transfer
to low-data settings, and that the sample size required for reliable BDL
evaluation is substantially larger than practitioners typically assume
\citep{dehghani2021benchmark, raji2021everything, reuel2024betterbench}.
Our framework gives researchers a way to determine whether their
evaluation data is sufficient before drawing conclusions about method
superiority.

\section{Related Work}

The systematic evaluation of uncertainty quantification in deep learning has a
substantial history.
\citet{lakshminarayanan2017simple} proposed Deep Ensembles and demonstrated
their superiority over approximate Bayesian methods on standard regression and
classification benchmarks.
\citet{ovadia2019trust} provided the first large-scale evaluation of predictive
uncertainty under dataset shift, finding that Deep Ensembles consistently
outperform other methods as distribution shift increases.
\citet{seligmann2023beyond} extended this line of work across diverse
architectures and tasks from the WILDS benchmark, introducing signed calibration
metrics to distinguish overconfidence from underconfidence.
A feature shared by all these studies is that evaluation is conducted on
datasets with abundant training samples, typically thousands to tens of
thousands of examples, and the conclusions they draw are treated as general
truths about method performance.
Our work asks what happens when this abundance cannot be assumed.

The theoretical foundations for evaluating probabilistic forecasts rest on
proper scoring rules \citep{gneiting2007strictly}.
The Continuous Ranked Probability Score and Negative Log-Likelihood are both
strictly proper, meaning a forecaster maximizes their expected score only by
reporting their true predictive distribution.
In practice, several studies have observed that these metrics yield inconsistent
rankings across methods and datasets
\citep{ovadia2019trust, seligmann2023beyond}, but the source of this
inconsistency has not been attributed to finite-sample estimation variance.
We show that at small training set sizes, the variance of these estimators is
itself large enough to render pairwise method comparisons unreliable.

A parallel line of work examines the statistical reliability of evaluation
scores more broadly.
\citet{bouthillier2021accounting} showed that variance due to training
randomness is frequently ignored in benchmark comparisons, leading to misleading
conclusions about algorithmic superiority.
\citet{longjohn2025statistical} proposed bootstrapping and Bayesian hierarchical
models to quantify uncertainty in aggregate performance metrics for foundation
model benchmarks, focusing on accuracy-based metrics in large-data settings.
Our work differs from both in a way that matters for practice.
We study the scarcity of training data rather than test data or training
randomness alone, and in low-data regimes the same dataset serves simultaneously
as the source of model fit and the basis for metric estimation, so sampling
variance propagates through both at once.
This compounding effect is absent when evaluation data is plentiful, which is
why existing frameworks do not surface it.

\section{Evaluation Framework}
\label{sec:method}

\subsection{Methods and Metrics Under Comparison}

We evaluate six representative approaches to uncertainty quantification in
regression.
Maximum A Posteriori estimation serves as a deterministic baseline with
heteroscedastic variance estimated via negative log-likelihood training.
Monte Carlo Dropout \citep{gal2016dropout} approximates a Bayesian posterior
by applying dropout at test time and aggregating predictions across stochastic
forward passes.
Stochastic Weight Averaging Gaussian \citep{maddox2019simple} fits a Gaussian
distribution over network weights by collecting parameter snapshots during the
final phase of stochastic gradient descent.
Bayes By Backprop \citep{blundell2015weight} maintains explicit variational
posteriors over all network weights and optimizes the evidence lower bound.
Deep Ensembles \citep{lakshminarayanan2017simple} trains multiple independently
initialized networks and combines their predictions as a mixture of Gaussians.
Conformal Prediction
\citep{vovk2005algorithmic, angelopoulos2023conformal} constructs prediction
intervals with finite-sample coverage guarantees by calibrating residuals on a
held-out calibration set. To evaluate CP on distributional metrics, we fit a Gaussian to match the
conformal interval width. Since CP targets coverage rather than distributional
accuracy, CRPS and NLL results for CP should be interpreted as a Gaussianized
approximation, while PICP and MPIW remain its primary evaluation metrics.

We evaluate each method using four metrics.
The Continuous Ranked Probability Score \citep{gneiting2007strictly} measures
the integrated squared difference between the predicted cumulative distribution
function and the empirical distribution of the observation, and is our primary
metric because it jointly rewards calibration and sharpness without requiring
a parametric distributional assumption at evaluation time.
Negative Log-Likelihood evaluates the log density of observations under the
predictive distribution.
The Prediction Interval Coverage Probability measures the empirical fraction
of test observations falling within the nominal 90\% prediction interval.
The Mean Prediction Interval Width measures the average width of those
intervals.
Full implementation details are provided in
Appendix~\ref{app:implementation}.

\subsection{Bayesian Hierarchical Model for Metric Uncertainty}

Standard practice reports a single metric value per method, computed on one
draw of $n$ training samples.
This point estimate conflates the true performance of a method with the
particular training set that happened to be drawn.
When $n$ is small, different draws of the same size can yield substantially
different metric values for the same method, and comparing point estimates
across methods amounts to comparing two noisy realizations of an underlying
quantity, neither of which is well estimated.

To separate signal from noise, we treat the metric value for each method as a
random variable across independent draws of the training data.
Let $y_{m,i}$ denote the value of a given metric for method $m$ on data
realization $i$, where each realization corresponds to an independently drawn
training set of size $n$ from the same data pool.
We model the distribution of metric values as
\begin{align}
    y_{m,i} &\sim \mathcal{N}(\mu_m + \gamma_i,\, \sigma_m^2)
    \label{eq:bhm} \\
    \gamma_i &\sim \mathcal{N}(0,\, \sigma_\gamma^2) \notag \\
    \mu_m &\sim \mathcal{N}(\mu_0,\, \tau^2) \notag \\
    \mu_0 &\sim \mathcal{N}(0,\, 1), \quad
    \tau \sim \text{HalfNormal}(1), \quad
    \sigma_m \sim \text{HalfNormal}(1), \quad
    \sigma_\gamma \sim \text{HalfNormal}(1). \notag
\end{align}
Here $\mu_m$ is the latent true metric value for method $m$, $\sigma_m^2$
captures within-method variance, $\tau^2$ captures between-method variance, and
$\gamma_i$ is a realization-specific random intercept that accounts for the
paired structure across methods due to the shared training subsample within
each realization $i$.
Allowing method-specific variances $\sigma_m^2$ is essential because methods
differ substantially in metric variability, particularly at small $n$ (e.g.,
a 1.5x difference in empirical CRPS standard deviation at $n = 50$).
A shared variance specification would pool these differences and may affect
posterior comparisons.
We therefore use method-specific variances throughout.
We verify through posterior predictive checks that the method-specific Gaussian
likelihood provides adequate fit for CRPS, which is continuous and
approximately symmetric across realizations at fixed $n$
\citep{gneiting2007strictly}.
For NLL, which can exhibit heavier tails at small $n$ due to occasional
optimization failures, we treat the Gaussian likelihood as a working
approximation and report sensitivity results.
Full diagnostics, including posterior predictive checks for all metrics, are
provided in Appendix~\ref{app:diagnostics}.

For PICP, which counts covered test points out of $N_{\text{test}}$, we use a
Beta-Binomial model \citep{griffiths1973maximum}.
Let $k_{m,i}$ be the number of covered points for method $m$ on realization
$i$.
Then
\begin{equation}
    k_{m,i} \sim \text{Beta-Binomial}(N_{\text{test}}, \mu_m, \phi), \quad
    \text{logit}(\mu_m) \sim \mathcal{N}(\mu_0, \tau^2),
    \label{eq:betabinom}
\end{equation}
where $\phi > 0$ is a precision parameter.
This respects the discrete count nature of PICP and accommodates
overdispersion \citep{menssen2019prediction}.

We fit each model using four Markov chain Monte Carlo chains with 2000
iterations in Stan via the \texttt{brms} interface \citep{burkner2017brms}.
Convergence is assessed using $\hat{R} \leq 1.01$ and bulk effective sample
size above 400; all fitted models satisfy both criteria
(Appendix~\ref{app:diagnostics}).

Given posterior samples $\{\mu_m^{(s)}\}_{s=1}^S$ for each method $m$, we
define the posterior probability that method $A$ achieves a lower metric value
than method $B$ as
\begin{equation}
    P(A \prec B) = \frac{1}{S} \sum_{s=1}^{S}
    \mathbf{1}\!\left[\mu_A^{(s)} < \mu_B^{(s)}\right].
    \label{eq:posterior_prob}
\end{equation}
A value of $P(A \prec B) > 0.95$ indicates that the data strongly support $A$
outperforming $B$.
Values near 0.5 indicate that the comparison is inconclusive given the
available data.
Because both $\mu_A$ and $\mu_B$ are drawn from the same joint posterior fit
under the hierarchical model, this probability captures dependencies between
them.

This hierarchical formulation is useful for two reasons. First, at small $n$,
method-specific metric variances are themselves difficult to estimate reliably,
and partial pooling stabilizes these estimates without imposing a shared-variance
assumption. Second, posterior samples provide a coherent joint description of
$(\mu_A,\mu_B,\sigma_A,\sigma_B)$, allowing posterior ordering and predictive
MDD to be computed within the same framework.

\subsection{Minimum Detectable Difference}

Posterior ordering addresses whether the available data support a superiority
claim, but it does not by itself determine whether the observed gap would be
reliably detected in a new experiment at the same training size.
To answer that predictive question, we require coherent uncertainty propagation
for both the mean difference and the method-specific variability.
This is the main reason we work with the hierarchical posterior rather than a
stand-alone plug-in comparison: the predictive MDD is defined from the same
joint uncertainty structure used for posterior ordering.

Even with a fitted Bayesian hierarchical model, practitioners need a practical
criterion for experimental design: given $n$ training samples, how large must a
performance difference be to be reliably detected in a new experiment?

Following the predictive perspective on sample size determination
\citep{spiegelhalter1986}, we define a predictive Minimum Detectable Difference
(MDD) as a threshold on the true difference $\Delta$ such that
\[
P(\Delta_{\text{new}} > 0 \mid \Delta, \mathbf{y}) \ge \gamma,
\]
where $\Delta_{\text{new}}$ denotes the difference in observed metric values in
a future replication. Under a Gaussian approximation,
\[
\Delta_{\text{new}} \sim \mathcal{N}(\Delta, \sigma_{\text{pred}}^2), \quad
P(\Delta_{\text{new}} > 0) =
\Phi\!\left(\frac{\Delta}{\sigma_{\text{pred}}}\right),
\]
so solving $P(\Delta_{\text{new}} > 0) = \gamma$ yields
\begin{equation}
    \text{MDD}(n, \gamma) = z_\gamma \cdot \sigma_{\text{pred}}.
\end{equation}
Here $z_\gamma$ denotes the $\gamma$-quantile of the standard normal, and
\[
\sigma_{\text{pred}}^2 =
\operatorname{Var}(\mu_A - \mu_B \mid \mathbf{y}) + \sigma_A^2 + \sigma_B^2.
\]
The first term captures posterior uncertainty in the mean difference, while
$\sigma_A^2 + \sigma_B^2$ accounts for method-specific variability in new
observations. We adopt $\gamma = 0.80$
\citep{cohen1988statistical}. A performance gap smaller than
$\text{MDD}(n, 0.80)$ is therefore unlikely to be reliably detected in a new
experiment at that sample size.

\section{Experimental Protocol}

We evaluate on one synthetic and four real-world regression datasets.
The synthetic dataset is generated with $d = 8$ dimensional inputs and
heteroscedastic noise $\sigma(x) = 0.3 + 0.5|x_1|$, requiring each method
to learn both the conditional mean and the input-dependent predictive variance.
We generate 1200 samples in total; the final 30\% form a fixed test set used
consistently across all experiments, and the remaining 840 samples serve as
the training pool.
The four real-world datasets are Energy Efficiency
\citep{tsanas2012accurate}, Concrete Compressive Strength
\citep{yeh1998modeling}, Kin8nm \citep{ghahramani1996kin}, and Yacht
Hydrodynamics \citep{gerritsma1981geometry}, covering a range of pool sizes
and input dimensions.
Dataset summaries are provided in Appendix~\ref{app:datasets}.

For each combination of method and training size
$n \in \{30, 50, 100, 200, 500\}$, we independently draw $R = 50$ random
subsamples from the training pool without replacement.
Each subsample defines a distinct data realization on which the method is
trained from scratch and evaluated against the fixed test set, yielding 50
metric observations per cell as input to the Bayesian hierarchical model in
Section~\ref{sec:method}.
The choice of $R = 50$ balances the statistical stability of the hierarchical
model against computational cost; we verify that posterior estimates are
stable under this choice in Appendix~\ref{app:diagnostics}.
All stochastic elements use fixed random seeds indexed by realization number
to ensure reproducibility.

All six methods share a common two-hidden-layer MLP architecture with 64 units
per layer and ReLU activations.
Full training details including optimizer settings, learning rates, and
method-specific configurations are provided in
Appendix~\ref{app:implementation}.

\section{Results}
\label{sec:results}

\textcolor{black}{This section proceeds in four steps. Section~5.1 quantifies
metric fluctuation across repeated low-data realizations. Section~5.2 examines
whether this variability makes pairwise comparisons inconclusive. Section~5.3
studies predictive detectability through the Minimum Detectable Difference.
Section~5.4 tests whether these patterns persist on four real-world datasets.}

\subsection{Metric Stability Under Data Scarcity}

Table~\ref{tab:crps} reports mean and standard deviation of CRPS across
$R = 50$ independent data realizations. Estimation variance decreases
monotonically with $n$ for all methods, confirming that metric reliability
improves with training set size. Importantly, the standard deviation of CRPS
remains substantial throughout the range studied: at $n = 50$, the standard
deviation is 0.015 for MCD, 0.017 for Deep Ensembles, and 0.023 for MAP.
These magnitudes are non-negligible relative to the performance differences
between methods, directly motivating the hierarchical analysis that follows.

\begin{figure}[ht]
    \centering
    \includegraphics[width=0.8\linewidth]{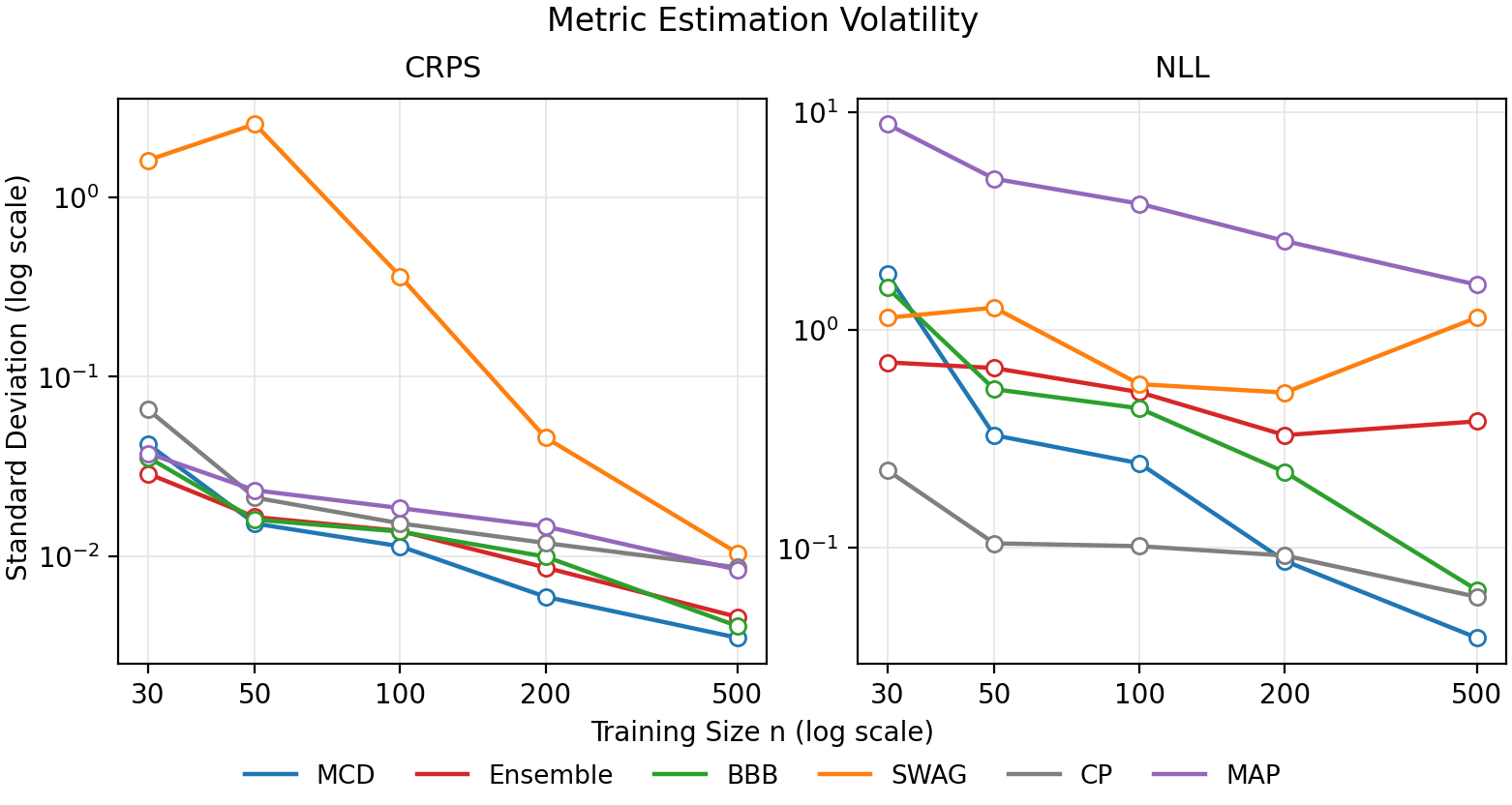}
    \caption{Standard deviation of CRPS and NLL across $R = 50$ data
    realizations, plotted on a log-log scale. All methods exhibit decreasing
    estimation variance with $n$, at rates that vary across methods
    (Table~\ref{tab:powerlaw}). SWAG exhibits anomalously high variance at
    small $n$ due to convergence failures.}
    \label{fig:volatility}
\end{figure}

Figure~\ref{fig:volatility} shows the standard deviation of CRPS and NLL across
realizations on a log-log scale. CRPS is substantially more stable than NLL at
all training sizes and across all methods. For NLL, variance is particularly
large for MAP, whose predicted variance collapses during training and produces
extreme NLL values at small $n$. SWAG exhibits anomalously high variance at
small $n$ due to stochastic gradient descent convergence failures; we treat
such failures as an implementation constraint and exclude failed runs from
metric computation, benchmarking the uncertainty quantification quality of
successful SWAG fits rather than its convergence reliability. Convergence rates
across realizations are reported in Appendix~\ref{app:swag}. A power-law fit
$\text{Var}[\widehat{\text{CRPS}}_n] \propto n^{-\alpha}$ provides a
descriptive summary of the rate of decrease, with estimated exponents ranging
from 0.49 for MAP to 0.83 for MCD
(Table~\ref{tab:powerlaw}). This variation in convergence rates arises because
algorithm variance (due to training randomness) compounds data variance (due
to sampling) differently across methods, as illustrated in
Figure~\ref{fig:variance_decomp}; methods with smaller $\alpha$ require
disproportionately more data to achieve reliable metric estimates.

Coverage-based metrics exhibit similar fragility. At $n = 50$, the standard
deviation of PICP is 0.047 for MCD and 0.169 for SWAG, meaning that a
single-replication coverage estimate at this sample size is unreliable
regardless of the method. Among proper scoring rules, CRPS provides
substantially more stable small-sample estimates than NLL, and we use it as
the primary metric in the analysis that follows. Cross-metric rank consistency increases with $n$ but remains below 1.0 even at
$n=500$ (Appendix~C.3), confirming that CRPS and Interval Score are not
interchangeable.

\begin{table}[ht]
\caption{Mean CRPS with standard deviation in parentheses across $R = 50$ data realizations. Lower is better. SWAG results at $n = 30, 50$ include only converged runs (Appendix~A.5). Best mean in each column is shown in bold.}
\label{tab:crps}
\centering
\small
\setlength{\tabcolsep}{6pt}
\renewcommand{\arraystretch}{1.12}
\begin{tabular}{lccccc}
\toprule
Method & $n=30$ & $n=50$ & $n=100$ & $n=200$ & $n=500$ \\
\midrule
MAP      & 0.310 (0.037) & 0.277 (0.023) & 0.261 (0.019) & 0.247 (0.015) & 0.216 (0.008) \\
MCD      & 0.240 (0.042) & \textbf{0.192} (0.015) & \textbf{0.178} (0.011) & \textbf{0.161} (0.006) & \textbf{0.149} (0.004) \\
Ensemble & \textbf{0.223} (0.029) & 0.206 (0.017) & 0.194 (0.014) & 0.186 (0.009) & 0.170 (0.005) \\
\rowcolor{swagbg}
SWAG     & 1.399 (1.607) & 1.287 (2.553) & 0.424 (0.362) & 0.190 (0.046) & 0.175 (0.010) \\
BBB      & 0.236 (0.036) & 0.202 (0.016) & 0.194 (0.014) & 0.179 (0.010) & 0.162 (0.004) \\
CP       & 0.314 (0.066) & 0.240 (0.021) & 0.218 (0.015) & 0.208 (0.012) & 0.185 (0.009) \\
\bottomrule
\end{tabular}
\end{table}

\subsection{Posterior Probabilities of Method Ranking}
\label{sec:ranking}

\begin{figure}[t]
    \centering
    \includegraphics[width=0.98\linewidth]{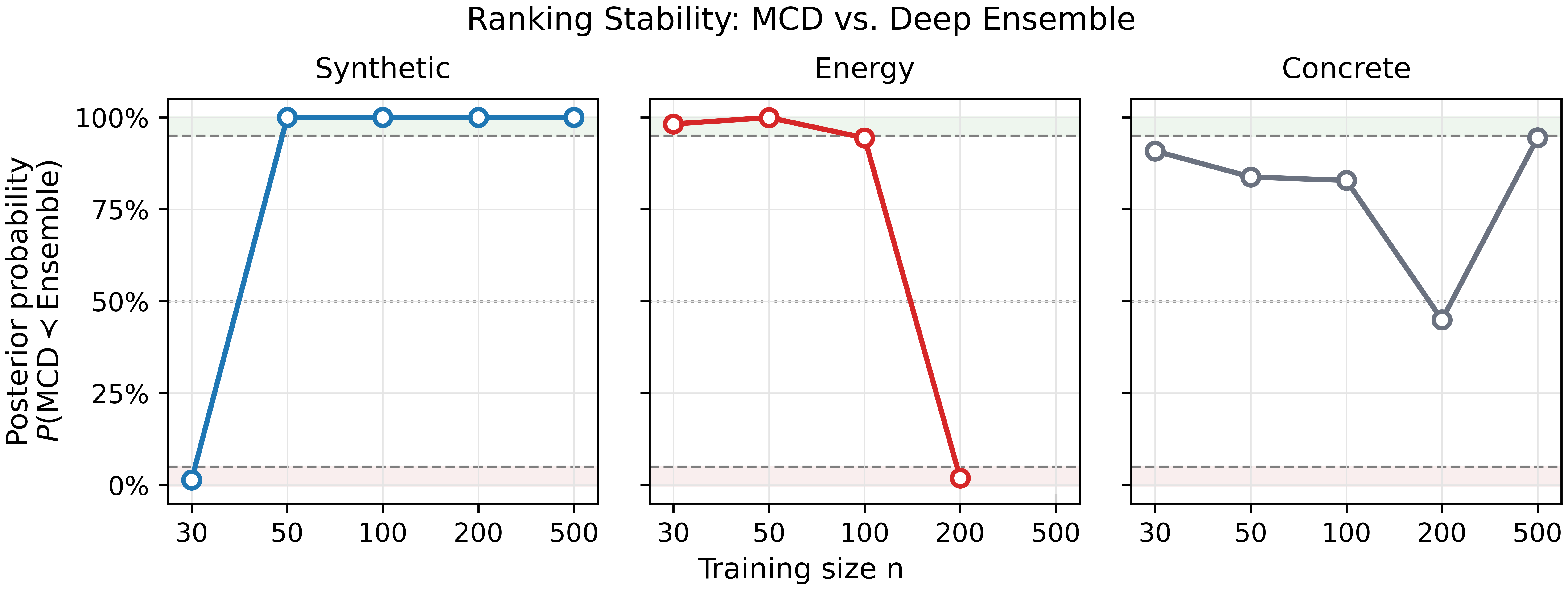}
\caption{Posterior probability $P(\mathrm{MCD} \prec \mathrm{Ensemble})$ on
CRPS across training sizes for three representative datasets. Synthetic and
Energy exhibit ranking reversals, while Concrete remains weaker and nearly
inconclusive at $n=200$. Dashed lines at $0.05$ and $0.95$ indicate strong
posterior support, and the dotted line at $0.50$ indicates an inconclusive
comparison.}
    \label{fig:posterior_probs}
\end{figure}

Figure~\ref{fig:posterior_probs} shows the posterior probability
$P(\text{MCD} \prec \text{Ensemble})$ as a function of $n$ on the synthetic
dataset. At $n = 30$, $P = 0.014$, indicating Deep Ensemble superiority; the
ranking reverses at $n = 50$ ($P = 1.000$) and remains stable thereafter. On
the Energy dataset, a more striking pattern emerges: MCD dominates at small
$n$ ($P = 1.000$ at $n = 50$) but the ranking reverses again at $n = 200$
($P = 0.019$), where Deep Ensembles regain superiority. These patterns
illustrate that method rankings are not stable in training set size: a single
reversal can occur early and stabilize, or multiple reversals can emerge as
$n$ grows.

Posterior probabilities of each method against the MAP
baseline are provided in Appendix~\ref{all:pair}.

\begin{figure}[t]
    \centering
    \includegraphics[width=0.9\linewidth]{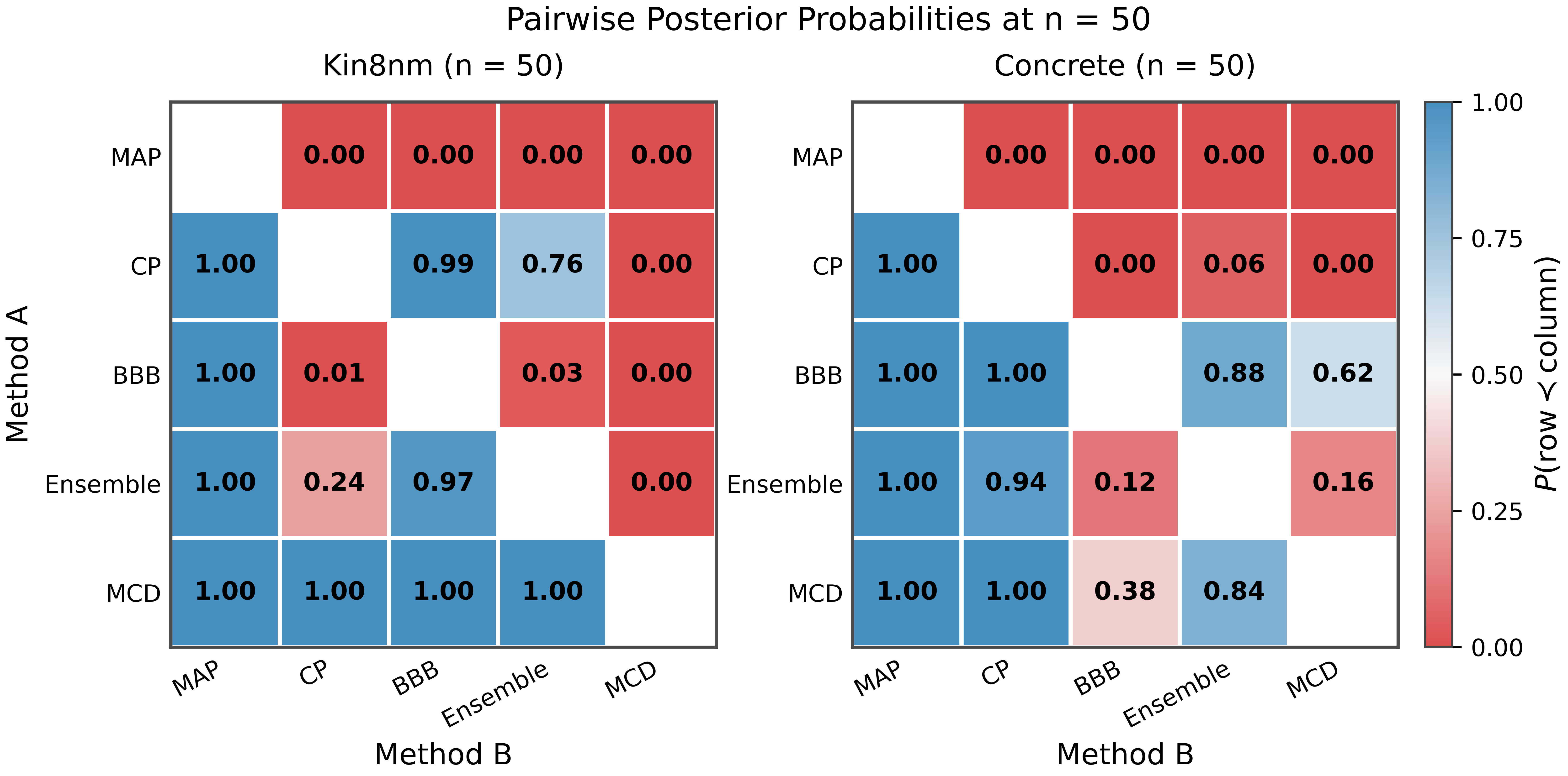}
    \caption{Pairwise posterior probability $P(\text{row} \prec
    \text{column})$ on CRPS for Kin8nm (left) and Concrete (right) at
    $n = 50$. Values $> 0.95$ (dark blue) indicate that the row method
    reliably outperforms the column method; values $< 0.05$ (dark red)
    indicate the opposite; values near $0.5$ (white) indicate inconclusive
    comparisons. The stark contrast between datasets illustrates that method
    rankings are dataset-dependent and cannot be generalized from a single
    benchmark.}
    \label{fig:pairwise_heatmap_n50}
\end{figure}

The pattern in Figure~\ref{fig:posterior_probs} extends beyond the
MCD--Ensemble comparison. Figure~\ref{fig:pairwise_heatmap_n50} shows the full
pairwise posterior probability matrix for two representative datasets at
$n = 50$, revealing that the determinism of rankings varies dramatically across
datasets. On Concrete (top), most comparisons are inconclusive (white
off-diagonals), while on Kin8nm (bottom), the ranking is nearly deterministic
(blue/red saturation). This contrast reinforces our central claim: method
rankings are not portable across datasets under data scarcity. Additional
comparisons are provided in Appendix~\ref{pair:post}.

\subsection{Minimum Detectable Difference}

\begin{figure}[t]
    \centering
    \includegraphics[width=0.7\linewidth]{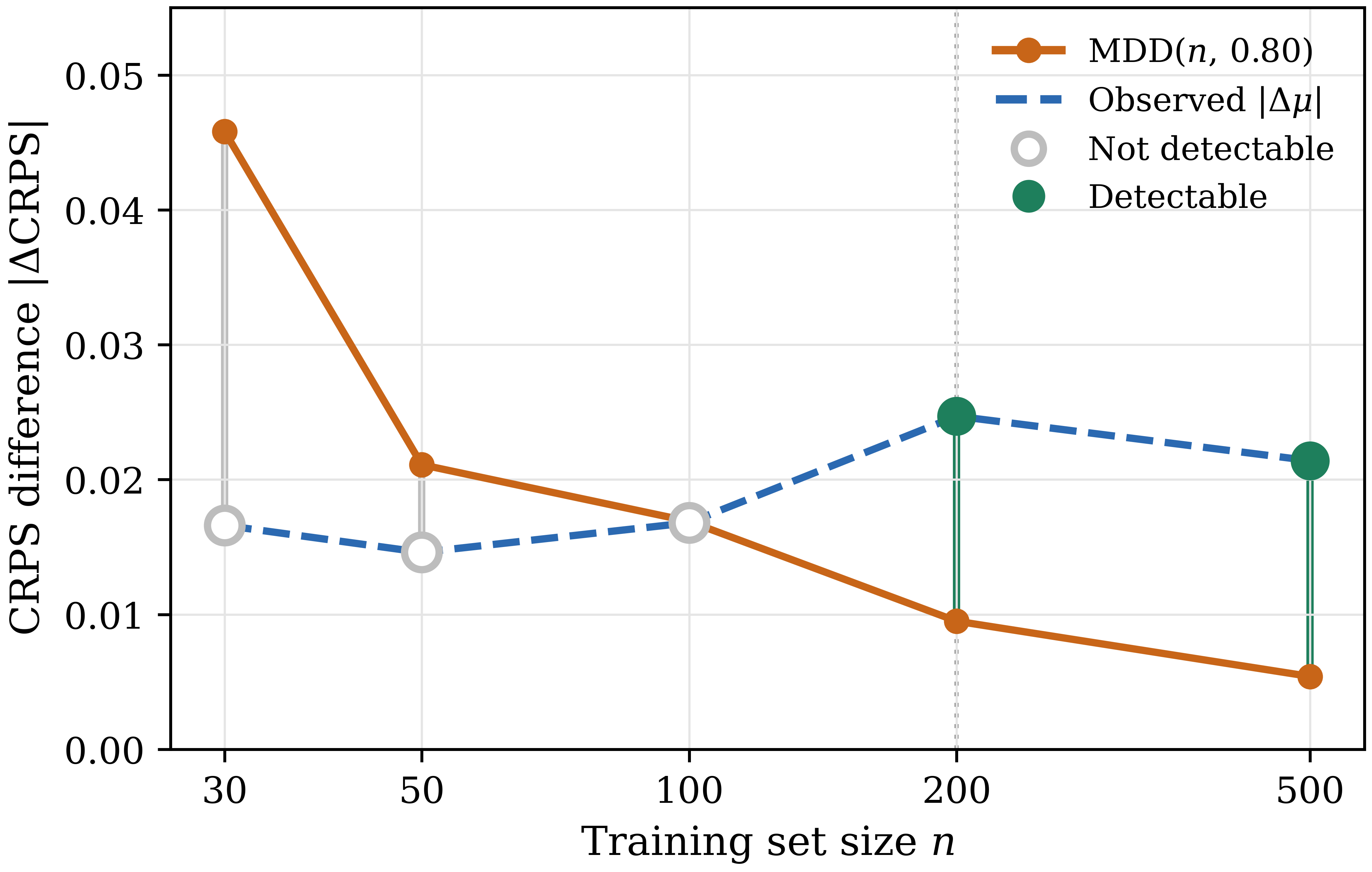}
    \caption{Predictive Minimum Detectable Difference for MCD vs.\ Deep
    Ensembles on CRPS. The red line is $\text{MDD}(n, 0.80)$, the smallest
    difference detectable in a new experiment with 80\% probability. The blue
    dashed line shows the observed mean difference
    $|\mu_{\text{MCD}} - \mu_{\text{Ensemble}}|$. Pink indicates undetectable
    regions, and green indicates detectable regions.}
    \label{fig:mdd}
\end{figure}

While Section~5.2 quantifies our certainty about population means
($P(\mu_{\text{MCD}} < \mu_{\text{Ensemble}}) = 1.000$ at $n = 50$),
practitioners often need a different answer: would a new experiment reliably
detect this difference? Figure~\ref{fig:mdd} shows the predictive MDD for MCD
versus Ensemble on CRPS.

At $n = 50$, the MDD is 0.021, larger than the observed mean difference of
0.015. This means that while we are certain about the population ordering, a
new experiment would detect the difference with only 72\% probability.
Achieving 80\% predictive power requires $n \geq 200$.

Two practical lessons follow. First, standard point-estimate comparisons
conflate population inference with predictive replication, leading to
overconfidence in small-sample findings. Second, CRPS requires substantially
fewer training samples to detect real performance differences than
coverage-based metrics (Appendix~\ref{app:real}), confirming that metric choice
matters not only for accuracy but for statistical power.

\subsection{Validation on Real Datasets}
\label{sec:real}

Table~\ref{tab:pairwise} shows that the ranking instability observed on the
synthetic dataset replicates across all four real regression datasets, but with
markedly different patterns. Three findings emerge that go beyond the synthetic
analysis.

First, the direction of the ranking varies by dataset in ways that are not
predictable from synthetic results alone. On Energy, MCD dominates at small
$n$ ($P = 1.000$ at $n = 50$) but the ranking reverses at $n = 200$
($P = 0.019$), where Deep Ensembles become the superior method. On Yacht,
Deep Ensembles dominate throughout ($P = 0.009$ at $n = 50$, $P = 0.000$ at
$n = 100$). On Concrete, the comparison remains genuinely uncertain across most
training sizes, with $P = 0.838$ at $n = 50$ and $P = 0.450$ at $n = 200$,
meaning that even two hundred training samples are insufficient to establish a
reliable winner on this dataset. At $n = 500$, the probability reaches
$0.945$, just below the $0.95$ threshold.

Second, MAP remains overconfident across all datasets with PICP near zero,
while Conformal Prediction maintains coverage closest to the nominal level on
some datasets (Kin8nm, Energy) but degrades substantially on others
(Concrete, Yacht) as $n$ increases, demonstrating that the empirical
performance of conformal prediction varies across datasets in our study.
Complete CRPS, PICP, and pairwise probability results for all datasets are
reported in Appendix~\ref{app:real}.

Third, these results suggest a practical principle for low-data evaluation.
Point-estimate leaderboards alone should not be treated as sufficient evidence
of method superiority when training data is limited, since the support for a
claimed ranking can remain weak and dataset-dependent even as $n$ grows.
Instead, practitioners should pair a stable primary metric such as CRPS with an
uncertainty-aware comparison statistic such as posterior superiority
probability, and use a detectability diagnostic such as the MDD curve to
determine whether the available training size is large enough to support a
winner claim at all. When feasible, conclusions should also be checked across
more than one dataset, since rankings that appear favorable in one setting can
reverse in another as training size changes.

\begin{table}[t]
\caption{Posterior probability $P(\mathrm{MCD} \prec \mathrm{Ensemble})$ on CRPS across datasets and training sizes. Values near $0$ indicate Deep Ensemble superiority, values near $1$ indicate MCD superiority, and values near $0.5$ indicate inconclusive comparisons. --- denotes conditions unavailable due to dataset size.}
\label{tab:pairwise}
\centering
\small
\setlength{\tabcolsep}{7pt}
\renewcommand{\arraystretch}{1.12}
\begin{tabular}{lccccc}
\toprule
Dataset   & $n=30$ & $n=50$ & $n=100$ & $n=200$ & $n=500$ \\
\midrule
Synthetic & \cellcolor{lowred}0.01 & \cellcolor{highblue}1.00 & \cellcolor{highblue}1.00 & \cellcolor{highblue}1.00 & \cellcolor{highblue}1.00 \\
Energy    & \cellcolor{highblue}0.98 & \cellcolor{highblue}1.00 & 0.94 & \cellcolor{lowred}0.02 & {\color{dashgray}---} \\
Concrete  & 0.91 & 0.84 & 0.83 & \cellcolor{midgray}0.45 & 0.95 \\
Kin8nm    & \cellcolor{highblue}0.99 & \cellcolor{highblue}1.00 & \cellcolor{highblue}1.00 & \cellcolor{highblue}1.00 & \cellcolor{highblue}1.00 \\
Yacht     & 0.13 & \cellcolor{lowred}0.01 & \cellcolor{lowred}0.00 & {\color{dashgray}---} & {\color{dashgray}---} \\
\bottomrule
\end{tabular}
\end{table}

\section{Conclusion}
\label{sec:conclusion}

This paper identifies a blind spot in current Bayesian deep learning
evaluation practice: when training data is limited, the metrics used to rank
methods are themselves unreliable, and the rankings themselves can reverse as
training size grows in ways that point-estimate comparisons cannot detect. On
the synthetic dataset, the posterior probability that Monte Carlo Dropout
outperforms Deep Ensembles on CRPS is $0.014$ at $n = 30$ and $1.000$ at
$n = 50$, a reversal that is invisible from mean CRPS values alone. On
Concrete, the same comparison remains statistically inconclusive at $n = 200$
($P = 0.450$), illustrating that no single sample size threshold separates
reliable from unreliable comparisons across datasets.

Our Bayesian hierarchical model framework addresses this gap by treating
evaluation metrics as random variables rather than fixed quantities, enabling
posterior inference over pairwise method comparisons. The MDD curve complements
this by translating posterior uncertainty into a practical sample-size
diagnostic. Unlike standard power analysis that considers only sampling
variability, our predictive MDD accounts for both mean uncertainty and
method-specific variance ($\sigma_m^2$), answering: ``Given $n$ training
samples, how large must a performance difference be to be reliably detected in
a new experiment?'' For the regression tasks studied here, detecting a true
CRPS difference between leading methods requires $n \geq 200$ to achieve 80\%
predictive power, confirming that metric choice matters not only for accuracy
but for statistical power.

Three empirical findings stand out. First, standard UQ metrics decrease in
variance with $n$ at rates that vary substantially across methods, with
power-law exponents ranging from 0.49 to 0.83, because model training
randomness compounds sampling variance in ways that practitioners often
overlook. Second, method rankings are not stable across datasets: Bayes By
Backprop outperforms Monte Carlo Dropout on the Concrete dataset at
$n \geq 100$, reversing the synthetic finding, while Deep Ensembles and Monte
Carlo Dropout exchange ranks on Energy at $n = 200$. Third, coverage-based
metrics remain unreliable at small training sizes, and among proper scoring
rules, CRPS provides substantially more stable small-sample estimates than
Negative Log-Likelihood. Conformal Prediction further illustrates this broader
instability: although it maintains near-nominal coverage on some datasets
(Energy, Kin8nm), its coverage degrades substantially on others
(Concrete, Yacht), showing that finite-sample coverage behavior does not
transfer uniformly across datasets.

Several limitations bound the scope of these conclusions. Our study focuses on
regression tasks with moderately sized neural networks; whether the same
patterns hold for classification, larger architectures, or spatially
structured data remains an open question. The Bayesian hierarchical model
assumes exchangeability across data realizations, which may be violated when
training sets are drawn from non-stationary distributions. Future work could
extend the framework to joint variation across training size, architecture, and
data modality, and examine whether the compounding variance phenomenon
persists in larger models trained with modern optimizers.

\bibliographystyle{plainnat}
\bibliography{ref}

\appendix

\section{Experimental Details}
\label{app:details}

\subsection{Posterior Distribution under Known Hyperparameters}

Conditioning on $\{\sigma_m\}$ and $\tau$, the posterior distribution
of $\mu_m$ given observations $y_{m,1}, \ldots, y_{m,R}$ is Gaussian
with mean and variance:
\begin{align}
V_m &= \left(\frac{R}{\sigma_m^2} + \frac{1}{\tau^2}\right)^{-1} \tag{A.1}\\
\hat{\mu}_m &= V_m \left(\frac{R\bar{y}_m}{\sigma_m^2} +
\frac{\mu_0}{\tau^2}\right) \tag{A.2}
\end{align}
where $\bar{y}_m = R^{-1}\sum_{i=1}^R y_{m,i}$ is the sample mean
and $\sigma_m^2$ is the within-method variance specific to method $m$.
This can be written as a shrinkage estimator:
\begin{equation}\tag{A.3}
\hat{\mu}_m = \lambda_m \bar{y}_m + (1 - \lambda_m)\mu_0, \qquad
\lambda_m = \frac{R/\sigma_m^2}{R/\sigma_m^2 + 1/\tau^2} \in [0,1]
\end{equation}
When $R$ is small relative to $\sigma_m^2/\tau^2$, $\lambda_m \to 0$
and the posterior shrinks toward the population mean $\mu_0$. When
$R$ is large, $\lambda_m \to 1$ and the posterior concentrates around
the sample mean. Methods with larger within-method variance $\sigma_m^2$
receive stronger shrinkage toward the population mean, which is the
mechanism by which the model accounts for heterogeneous noise levels
across methods.

Under known hyperparameters, the posterior probability of ranking
has a closed form:
\begin{equation}\tag{A.4}
P(\mu_A < \mu_B \mid y) = \Phi\!\left(\frac{\hat{\mu}_B -
\hat{\mu}_A}{\sqrt{V_A + V_B}}\right)
\end{equation}
where $\Phi$ is the standard normal CDF. In practice, $\{\sigma_m\}$
and $\tau$ are unknown and must be estimated from the data, which is
why we use Markov chain Monte Carlo rather than this closed-form
expression.




\subsection{Empirical Scaling of the Minimum Detectable Difference}

The Minimum Detectable Difference is defined as $\mathrm{MDD}(n) = z_\gamma \cdot \hat{\sigma}_\Delta(n)$ (see Section 3.3), where $z_\gamma$ is the $\gamma$-quantile of the standard normal distribution (we use $\gamma = 0.80$, giving $z_\gamma = 1.28$) and $\hat{\sigma}_\Delta(n)$ is the posterior standard deviation of $\mu_A - \mu_B$.

This definition follows directly from the posterior predictive distribution. Under the normal approximation to the posterior (verified via posterior predictive checks in Appendix~B), the predictive distribution for the difference in a new experiment is $\Delta_{\text{new}} \sim \mathcal{N}(\mu_A - \mu_B,\; \sigma_{\text{pred}}^2)$, where $\sigma_{\text{pred}}^2 = \operatorname{Var}(\mu_A - \mu_B \mid \mathbf{y}) + \sigma_A^2 + \sigma_B^2$. The probability that a new experiment detects Method A outperforming Method B is $P(\Delta_{\text{new}} < 0) = \Phi(-(\mu_A - \mu_B)/\sigma_{\text{pred}})$. Solving $P(\Delta_{\text{new}} < 0) \geq \gamma$ for $|\mu_A - \mu_B|$ yields $\mathrm{MDD} = z_\gamma \cdot \sigma_{\text{pred}}$, which is the closed-form expression used throughout the paper.

If we empirically observe that metric variance follows $\widehat{\mathrm{Var}}[\mathrm{CRPS}_n] \approx C n^{-\alpha}$ (Table~7), then the posterior standard deviation scales as $\hat{\sigma}_\Delta(n) \propto n^{-\alpha/2}$, giving:
\[
\mathrm{MDD}(n) \propto n^{-\alpha/2}
\]
This implies that methods with smaller $\alpha$ (e.g., MAP, $\alpha = 0.49$) require substantially more data to achieve the same detection power than methods with larger $\alpha$ (e.g., MCD, $\alpha = 0.83$). We emphasize that this scaling is an empirical description, not a theoretical guarantee.

\subsection{Datasets}
\label{app:datasets}

Table~\ref{tab:datasets} summarizes the five datasets used in our
experiments.
For each dataset, the pool size is computed after reserving a fixed
30\% test split.
Training subsamples of size $n$ are drawn from the pool without
replacement.

\begin{table}[htbp]
    \caption{Summary of datasets used for evaluation.}
    \label{tab:datasets}
    \centering
    \small
    \begin{tabular}{lrrrr}
        \toprule
        Dataset & Total & Features & Pool & $n$ levels \\
        \midrule
        Synthetic           & 1200 & 8 & 840  & 30, 50, 100, 200, 500 \\
        Energy Efficiency   & 768  & 8 & 538  & 30, 50, 100, 200 \\
        Concrete Strength   & 1030 & 8 & 721  & 30, 50, 100, 200, 500 \\
        Kin8nm              & 8192 & 8 & 5734 & 30, 50, 100, 200, 500 \\
        Yacht Hydrodynamics & 308  & 6 & 215  & 30, 50, 100 \\
        \bottomrule
    \end{tabular}
\end{table}

The synthetic dataset is generated with input $\mathbf{x} \sim
\mathcal{N}(0, I)$, $d = 8$, and fixed coefficient vector
$\mathbf{w} = [1.5, -2.0, 0.5, 1.0, -0.5, 0.3, -1.2, 0.8]^\top$.
All experiments use a global random seed of 42 for data generation.
The full dataset of 1200 samples is generated once and held fixed
across all experiments.
Training subsamples of size $n$ are drawn using seed $r + n$ for the
$r$-th data realization, ensuring reproducibility and preventing
overlap between realizations of the same $n$.

\subsection{Implementation Details}
\label{app:implementation}

All six methods share a common two-hidden-layer MLP architecture with
64 units per layer and ReLU activations.
The output layer produces two scalars corresponding to the predicted
mean $\mu$ and log variance $\log \sigma^2$.
Predicted variance is clamped to $[10^{-3}, 10^3]$ during training
for numerical stability.

MAP, MCD, and CP are trained for 500 epochs using the Adam optimizer \citep{kingma2015adam}
with learning rate $10^{-3}$ and weight decay $10^{-5}$.
BBB doubles the training budget to 1000 epochs to allow variational
convergence and omits weight decay, since the KL divergence term in
the evidence lower bound already provides regularization.
SWAG uses stochastic gradient descent with momentum 0.9, learning
rate $10^{-2}$, and cosine annealing, collecting weight snapshots
from epoch 300 onward.
For MCD, SWAG, and BBB, we draw $T = 50$ stochastic samples at test
time to form predictive distributions.
Conformal Prediction reserves 20\% of each training set as a
calibration split. All experiments were conducted on an NVIDIA T4 GPU, 
with a total compute of approximately 9 GPU-hours.

\subsection{SWAG Convergence}
\label{app:swag}

Stochastic Weight Averaging Gaussian relies on stochastic gradient descent to collect weight snapshots, and can fail to produce a meaningful posterior approximation when the training set is small. At small $n$, the gradient signal is weak and noisy, causing the SGD trajectory to wander rather than concentrate near a local minimum, which results in a weight posterior with artificially inflated variance. Even among converged runs, SWAG exhibits high variance at small $n$ due to sensitivity to initialization and SGD trajectory. Table~\ref{tab:swag_convergence} reports the fraction of $R = 50$ realizations in which SWAG successfully converged on each dataset. A run is considered failed if the loss becomes NaN or if fewer than two weight snapshots are collected during the averaging phase. Results reported for SWAG in the main paper are based on converged runs only. Convergence improves consistently with $n$ across all datasets, reaching near-perfect rates at $n \geq 200$.

\begin{table}[h]
\centering
\caption{SWAG convergence rate across $R = 50$ data realizations.
--- denotes training sizes not evaluated due to dataset pool size
constraints.}
\label{tab:swag_convergence}
\small
\begin{tabular}{lrrrrr}
\toprule
Dataset   & $n=30$ & $n=50$ & $n=100$ & $n=200$ & $n=500$ \\
\midrule
Synthetic & 0.58 & 0.88 & 0.98 & 1.00 & 1.00 \\
Energy    & 0.76 & 0.90 & 0.92 & 0.94 & ---  \\
Concrete  & 0.86 & 0.96 & 1.00 & 1.00 & 1.00 \\
Kin8nm    & 0.78 & 0.86 & 0.98 & 1.00 & 1.00 \\
Yacht     & 0.88 & 0.98 & 0.96 & ---  & ---  \\
\bottomrule
\end{tabular}
\end{table}

\subsection{BBB Hyperparameter Sensitivity}
\label{app:bbb}

The Kullback-Leibler divergence weight controls the strength of
regularization toward the prior in Bayes By Backprop. A weight
that is too large collapses the posterior toward the prior,
reducing predictive variance; a weight that is too small allows
the posterior to overfit the training data. Table~\ref{tab:bbb_kl}
reports mean CRPS at $n = 100$ on the synthetic dataset for three
values spanning two orders of magnitude. The results are stable
across the range, with all values falling within one standard
deviation of each other, confirming that the main conclusions are
not sensitive to this hyperparameter choice.

\begin{table}[h]
\centering
\caption{Mean CRPS at $n = 100$ for BBB under different
Kullback-Leibler divergence weights on the synthetic dataset
($R = 30$ repeats).}
\label{tab:bbb_kl}
\small
\begin{tabular}{lr}
\toprule
KL weight & Mean CRPS \\
\midrule
$10^{-5}$ & 0.187 \\
$10^{-4}$ & 0.186 \\
$10^{-3}$ & 0.182 \\
\bottomrule
\end{tabular}
\end{table}

\section{Model Diagnostics}
\label{app:diagnostics}

\paragraph{MCMC convergence.}
We assess convergence of the MCMC sampler using the potential scale
reduction factor $\hat{R}$ \citep{gelman1992inference} and the bulk
effective sample size.
All fitted models achieve $\hat{R} \leq 1.01$, and the minimum bulk
effective sample size across all fits is 521, well above the threshold
of 400.
Table~\ref{tab:rhat} reports the maximum $\hat{R}$ across all
parameters for each metric and training size.

\begin{table}[h]
    \caption{Maximum $\hat{R}$ across all parameters for each metric
    and training size. Values at or below 1.01 indicate satisfactory
    convergence.}
    \label{tab:rhat}
    \centering
    \small
    \begin{tabular}{lrrrrr}
        \toprule
        Metric & $n=30$ & $n=50$ & $n=100$ & $n=200$ & $n=500$ \\
        \midrule
        CRPS           & 1.00 & 1.01 & 1.01 & 1.00 & 1.00 \\
        PICP           & 1.01 & 1.00 & 1.00 & 1.00 & 1.01 \\
        Interval Score & 1.00 & 1.00 & 1.00 & 1.00 & 1.00 \\
        \bottomrule
    \end{tabular}
\end{table}

\paragraph{Likelihood specification.}
The Gaussian likelihood in Equation~\ref{eq:bhm} assumes that metric
values are approximately symmetric across data realizations at fixed
$n$.
For CRPS this assumption holds well across all methods and training
sizes.
For NLL, MAP exhibits heavy right skew at small $n$ due to sigma
collapse during training; we therefore treat the Gaussian likelihood
as an approximation for NLL at small $n$ and note that conclusions
based on NLL below $n = 100$ should be interpreted with caution.
SWAG at small $n$ is excluded from BHM fits where convergence rates
fall below 0.80, as the resulting metric distributions are not
representative of the method's true behavior.

\paragraph{Sensitivity to $R$.}
To verify that $R = 50$ realizations provides stable posterior
estimates, we refit the BHM using subsets of $R \in \{20, 30, 40,
50\}$ realizations drawn randomly from the full set of 50.
Posterior probability estimates change by less than 0.02 between
$R = 40$ and $R = 50$ across all method pairs and training sizes
examined, supporting our choice of $R = 50$.

\begin{figure}[h]
\centering
\includegraphics[width=\linewidth]{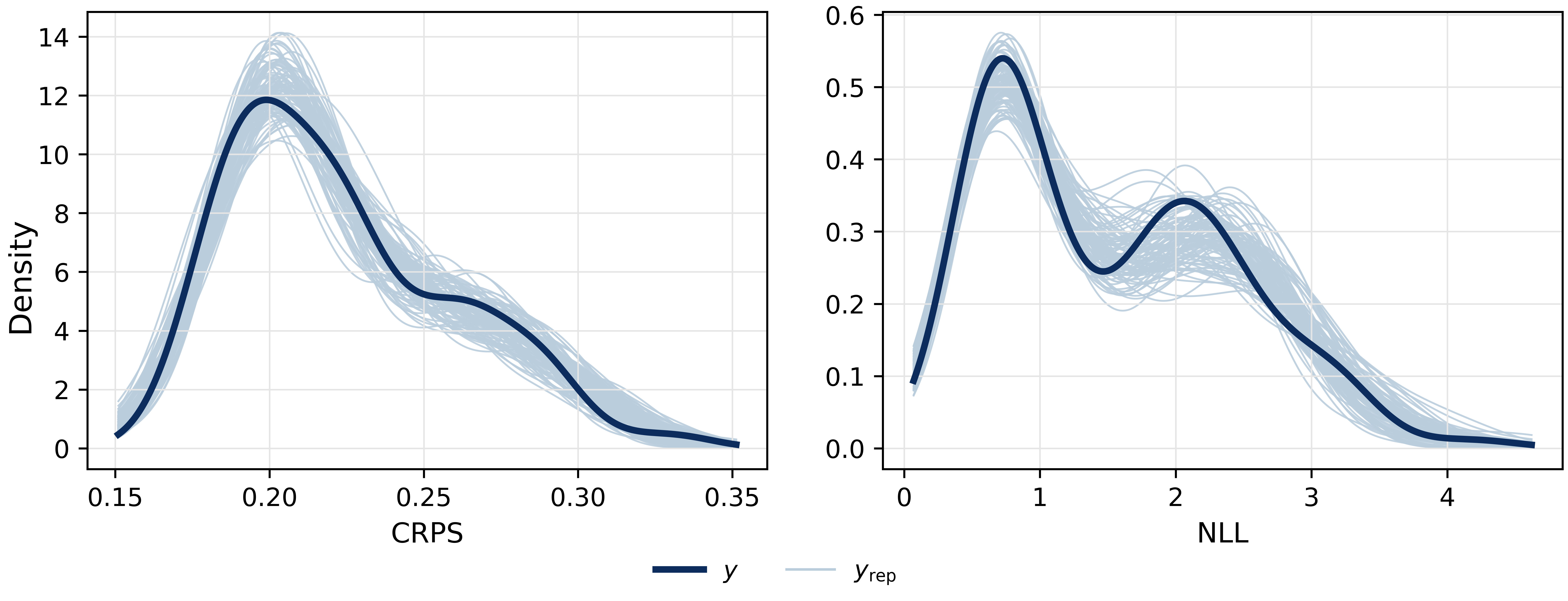}
\caption{Posterior predictive checks for CRPS (left) and NLL (right) at $n = 50$. Each light curve shows one draw of replicated data $y_{\text{rep}}$ from the posterior predictive distribution. The dark curve shows the observed data $y$. Close agreement indicates adequate model fit.}
\label{fig:ppc}
\end{figure}

\paragraph{Model fit.}
Figure~\ref{fig:ppc} shows posterior predictive checks for the Bayesian hierarchical model fit to CRPS and NLL at $n = 50$. For each metric, the replicated data $y_{\text{rep}}$ closely tracks the observed distribution $y$, indicating that the Gaussian likelihood provides adequate fit for both metrics under the method-specific variance specification.


\section{Complete Results}
\label{app:results}

\subsection{Pairwise Posterior Probabilities at Larger N}
\label{pair:post}

Figure~\ref{fig:pairwise_heatmap_n200} shows the full pairwise posterior probability matrix 
$P(\text{row} \prec \text{column})$ on CRPS at $n = 200$ for Kin8nm 
(left) and Concrete (right). At this sample size, Kin8nm exhibits a 
nearly deterministic ranking (dark blue/red), while Concrete remains 
largely inconclusive (white off-diagonals), reinforcing the main 
paper's finding that ranking reliability is dataset-dependent and 
cannot be generalized from a single benchmark. For $n = 50$ results, 
see Figure 4.
\begin{figure}[t]
    \centering
    \includegraphics[width=0.9\linewidth]{pic/fig3_heatmap_n200.png}
    \caption{Pairwise posterior probability $P(\text{row} \prec \text{column})$ on CRPS for Kin8nm (left) and Concrete (right) at $n = 200$. Values $>0.95$ (dark blue) indicate that the row method reliably outperforms the column method; values $<0.05$ (dark red) indicate the opposite; values near $0.5$ (white) indicate inconclusive comparisons. The stark contrast between datasets illustrates that method rankings are dataset-dependent and cannot be generalized from a single benchmark.}
    \label{fig:pairwise_heatmap_n200}
\end{figure}

\subsection{Power-law Exponents}
\label{app:powerlaw}

Table~\ref{tab:powerlaw} reports the estimated power-law exponent
$\alpha$ for each method on the synthetic dataset, fitted by ordinary
least squares on $\log \mathrm{Var}[\widehat{\mathrm{CRPS}}_n]$
versus $\log n$ across $n \in \{30, 50, 100, 200, 500\}$.
SWAG is excluded due to convergence failures at small $n$.

\begin{table}[h]
    \caption{Power-law exponent $\alpha$ for CRPS
    estimation variance. Larger $\alpha$ indicates faster
    variance reduction. Reported as descriptive summary only.}
    \label{tab:powerlaw}
    \centering
    \small
    \begin{tabular}{lr}
        \toprule
        Method & $\alpha$ \\
        \midrule
        MCD      & 0.83 \\
        BBB      & 0.68 \\
        CP       & 0.63 \\
        Ensemble & 0.62 \\
        MAP      & 0.49 \\
        \bottomrule
    \end{tabular}
\end{table}

\begin{figure}[h]
    \centering
    \includegraphics[width=0.7\linewidth]{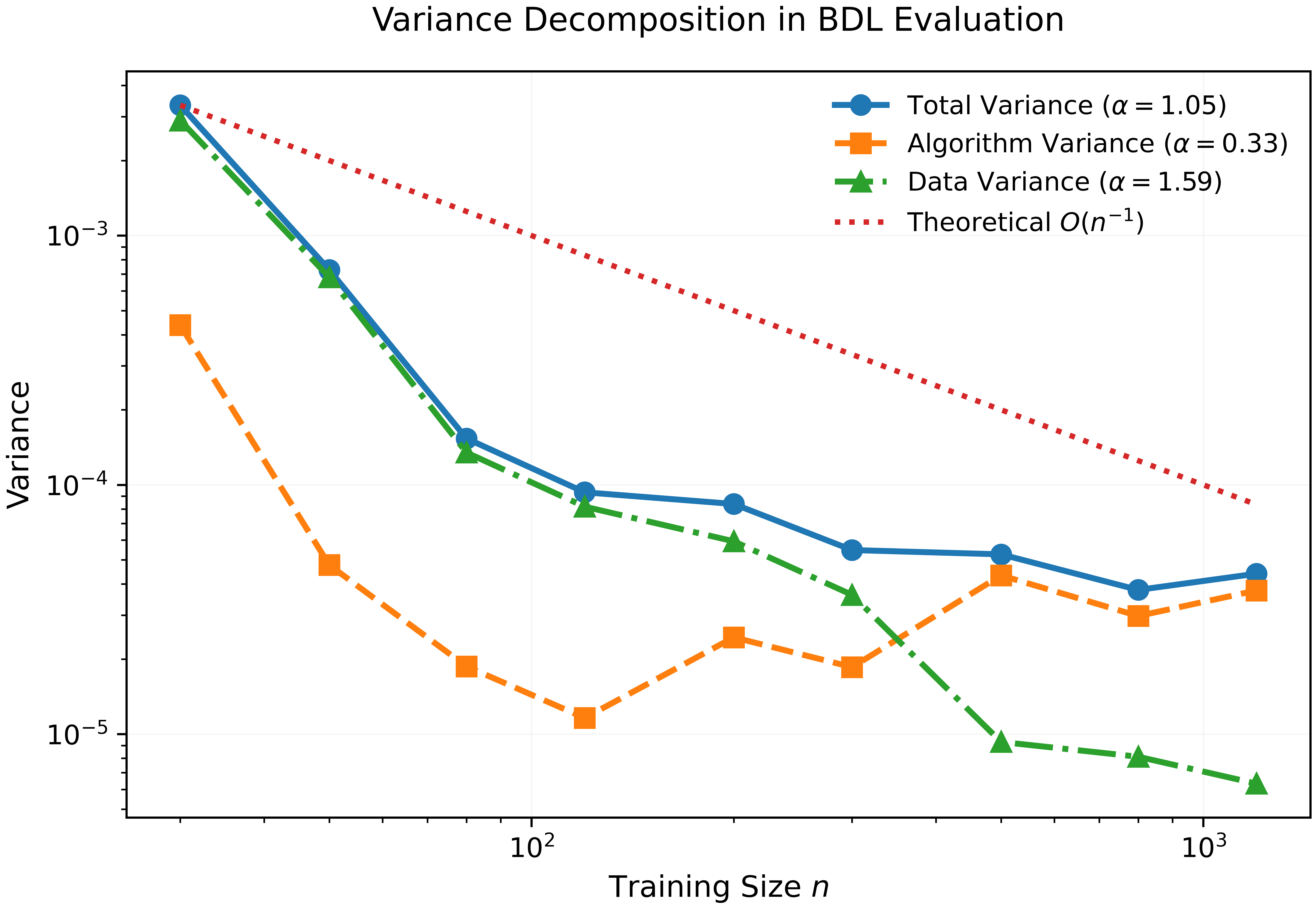}
    \caption{Variance decomposition of a neural network evaluation metric under repeated training on independent datasets of size $n$. Algorithm variance decreases substantially more slowly than data variance, explaining the range of power-law exponents observed in Table~\ref{tab:powerlaw}.}
    \label{fig:variance_decomp}
\end{figure}

\subsection{Full Synthetic Results}

Tables~\ref{tab:nll_full}--\ref{tab:mpiw_full} report mean $\pm$
standard deviation of NLL, PICP, and MPIW on the synthetic dataset. NLL values for MAP are computed using a sigma floor of $0.05$; the unclipped NLL for MAP exceeds $10^6$ at all training sizes due to sigma collapse under overconfident training.
NLL (clipped at $\sigma = 0.05$) across methods and
training sizes. Lower is better. MAP values are clipped due to sigma collapse; unclipped MAP NLL exceeds $10^6$ at all training
sizes. Deep Ensemble NLL is evaluated on the mixture predictive
distribution and does not decrease monotonically with $n$, a structural property of mixture-of-Gaussians representations documented in \citet{lakshminarayanan2017simple}. SWAG NLL increases at $n = 500$ due to posterior over dispersion under the stochastic weight averaging trajectory.

\begin{table}[h]
    \caption{NLL on the synthetic dataset. Lower is better.}
    \label{tab:nll_full}
    \centering
    \small
    \begin{tabular}{lrrrrr}
        \toprule
        Method & $n=30$ & $n=50$ & $n=100$ & $n=200$ & $n=500$ \\
        \midrule
        MAP
          & $36.57 \pm 8.79$ & $28.88 \pm 4.94$
          & $25.81 \pm 3.79$ & $22.87 \pm 2.56$ & $16.85 \pm 1.61$ \\
        MCD
          & $2.26 \pm 1.80$ & $0.96 \pm 0.33$
          & $0.72 \pm 0.24$ & $0.37 \pm 0.09$ & $0.14 \pm 0.04$ \\
        Ensemble
          & $2.25 \pm 0.71$ & $2.38 \pm 0.67$
          & $2.20 \pm 0.52$ & $2.00 \pm 0.33$ & $2.25 \pm 0.38$ \\
        SWAG
          & $2.18 \pm 1.14$ & $1.84 \pm 1.27$
          & $1.04 \pm 0.56$ & $0.69 \pm 0.52$
          & $2.18 \pm 1.14$ \\
        BBB
          & $3.63 \pm 1.56$ & $2.15 \pm 0.53$
          & $1.65 \pm 0.44$ & $0.99 \pm 0.22$ & $0.40 \pm 0.06$ \\
        CP
          & $0.89 \pm 0.23$ & $0.62 \pm 0.11$
          & $0.53 \pm 0.10$ & $0.48 \pm 0.09$ & $0.35 \pm 0.06$ \\
        \bottomrule
    \end{tabular}
\end{table}

\begin{table}[h]
    \caption{PICP on the synthetic dataset. Target value is 0.90.}
    \label{tab:picp_full}
    \centering
    \small
    \begin{tabular}{lrrrrr}
        \toprule
        Method & $n=30$ & $n=50$ & $n=100$ & $n=200$ & $n=500$ \\
        \midrule
        MAP
          & $0.001 \pm 0.001$ & $0.000 \pm 0.001$
          & $0.000 \pm 0.001$ & $0.001 \pm 0.001$ & $0.044 \pm 0.039$ \\
        MCD
          & $0.601 \pm 0.084$ & $0.689 \pm 0.047$
          & $0.719 \pm 0.038$ & $0.782 \pm 0.024$ & $0.842 \pm 0.013$ \\
        Ensemble
          & $0.595 \pm 0.052$ & $0.583 \pm 0.046$
          & $0.583 \pm 0.033$ & $0.599 \pm 0.031$ & $0.562 \pm 0.024$ \\
        SWAG
          & $0.915 \pm 0.154$ & $0.893 \pm 0.169$
          & $0.922 \pm 0.095$ & $0.796 \pm 0.133$ & $0.556 \pm 0.102$ \\
        BBB
          & $0.498 \pm 0.064$ & $0.569 \pm 0.039$
          & $0.603 \pm 0.037$ & $0.674 \pm 0.028$ & $0.765 \pm 0.016$ \\
        CP
          & $0.919 \pm 0.081$ & $0.928 \pm 0.054$
          & $0.882 \pm 0.063$ & $0.866 \pm 0.058$ & $0.869 \pm 0.032$ \\
        \bottomrule
    \end{tabular}
\end{table}

\begin{table}[h]
    \caption{MPIW on the synthetic dataset. Lower is better given
    adequate PICP.}
    \label{tab:mpiw_full}
    \centering
    \small
    \begin{tabular}{lrrrrr}
        \toprule
        Method & $n=30$ & $n=50$ & $n=100$ & $n=200$ & $n=500$ \\
        \midrule
        MAP
          & $0.000 \pm 0.000$ & $0.000 \pm 0.000$
          & $0.000 \pm 0.000$ & $0.000 \pm 0.000$ & $0.065 \pm 0.081$ \\
        MCD
          & $0.642 \pm 0.069$ & $0.653 \pm 0.039$
          & $0.653 \pm 0.032$ & $0.704 \pm 0.030$ & $0.775 \pm 0.019$ \\
        Ensemble
          & $0.637 \pm 0.062$ & $0.571 \pm 0.057$
          & $0.531 \pm 0.031$ & $0.525 \pm 0.028$ & $0.436 \pm 0.028$ \\
        SWAG
          & $7.978 \pm 4.835$ & $5.834 \pm 4.603$
          & $2.918 \pm 2.267$ & $0.964 \pm 0.598$ & $0.411 \pm 0.103$ \\
        BBB
          & $0.466 \pm 0.035$ & $0.491 \pm 0.024$
          & $0.515 \pm 0.022$ & $0.574 \pm 0.021$ & $0.671 \pm 0.013$ \\
        CP
          & $2.211 \pm 0.912$ & $1.675 \pm 0.414$
          & $1.265 \pm 0.300$ & $1.136 \pm 0.199$ & $0.993 \pm 0.092$ \\
        \bottomrule
    \end{tabular}
\end{table}

\subsection{Cross-metric Rank Consistency}
\label{app:consistency}

\begin{figure}[htbp]
    \centering
    \includegraphics[width=\linewidth]{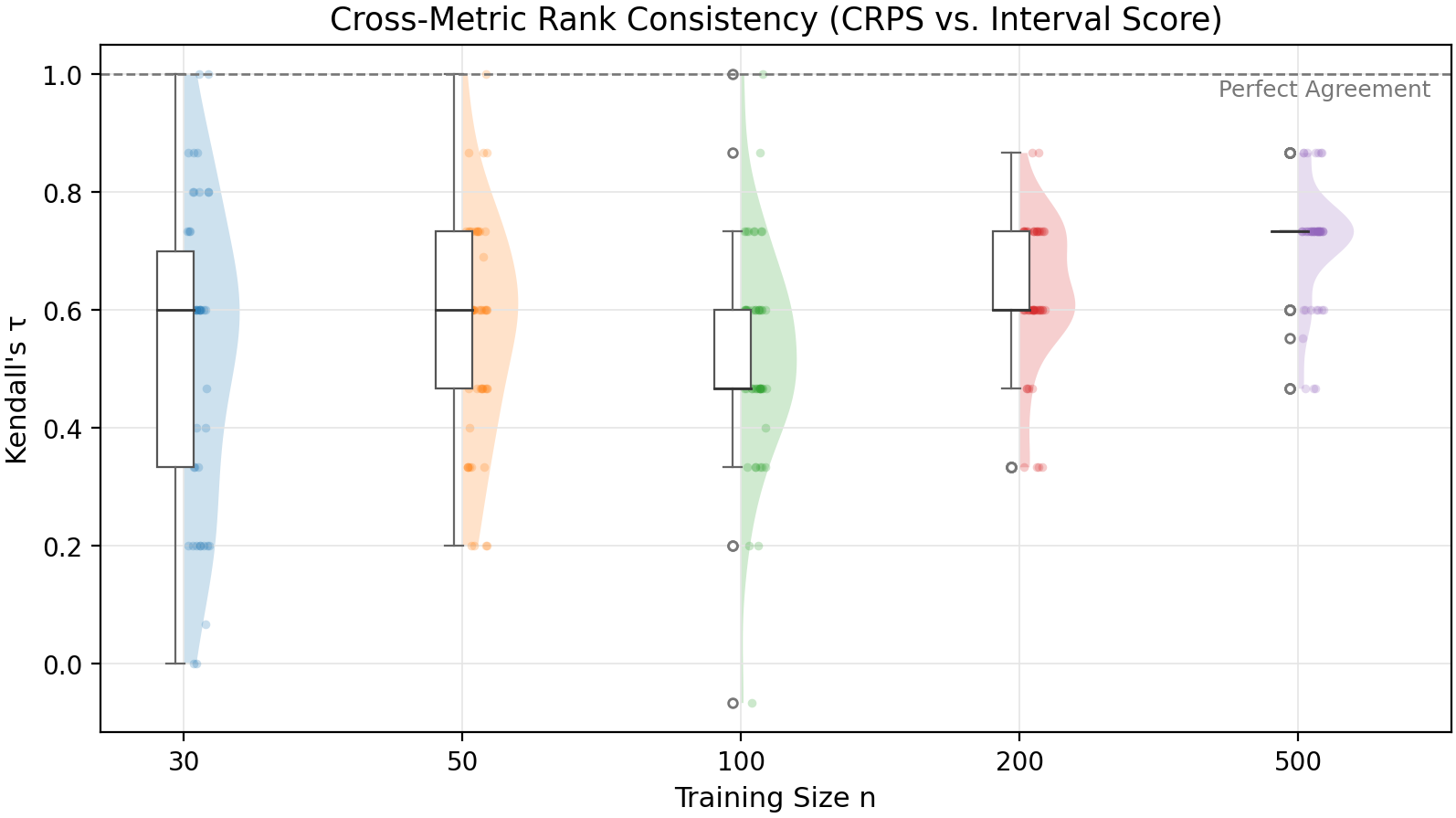}
    \caption{Distribution of Kendall $\tau$ between CRPS and Interval
    Score method rankings across $R = 50$ data realizations.
    Higher $\tau$ indicates greater agreement between the two metrics.
    The dashed line at $\tau = 1$ represents perfect agreement.
    Consistency increases with $n$ but never reaches $1.0$.}
    \label{fig:rank_consistency}
\end{figure}

Figure~\ref{fig:rank_consistency} shows the distribution of Kendall
$\tau$ between CRPS and Interval Score method rankings across $R = 50$
data realizations at each training size.
The median $\tau$ at $n = 30$ is approximately $0.58$, with an
interquartile range spanning $0.33$ to $0.68$.
At $n = 100$, the distribution widens and includes realizations where
$\tau$ approaches $-0.1$, meaning the two metrics occasionally produce
nearly reversed rankings.
The median $\tau$ increases to approximately $0.75$ at $n = 500$ but
does not approach $1.0$, confirming that CRPS and Interval Score are
not interchangeable even at larger training sizes.

\subsection{All Pairwise Posterior Probabilities}
\label{all:pair}

Table~\ref{tab:pairwise_map} reports the posterior probability
$P(\mathrm{method} \prec \mathrm{MAP})$ on CRPS for all methods and
real datasets, complementing the MCD versus Ensemble comparison in
the main text.

\begin{table}[h]
    \caption{Posterior probability $P(\mathrm{method} \prec
    \mathrm{MAP})$ on CRPS on real datasets. SWAG is excluded
    from the Bayesian hierarchical model at training sizes where
    its convergence rate falls below $1.00$; these cells are
    marked \textbf{---}.}
    \label{tab:pairwise_map}
    \centering
    \small
    \begin{tabular}{lrrrrr}
        \toprule
        Method & $n=30$ & $n=50$ & $n=100$ & $n=200$ & $n=500$ \\
        \midrule
        \multicolumn{6}{l}{\textit{Energy Efficiency}} \\
        MCD      & 1.000 & 1.000 & 1.000 & 1.000 & \textbf{---} \\
        Ensemble & 0.954 & 1.000 & 1.000 & 1.000 & \textbf{---} \\
        BBB      & 0.999 & 1.000 & 1.000 & 1.000 & \textbf{---} \\
        CP       & 0.001 & 0.000 & 0.058 & 0.919 & \textbf{---} \\
        \midrule
        \multicolumn{6}{l}{\textit{Concrete Compressive Strength}} \\
        MCD      & 1.000 & 1.000 & 1.000 & 1.000 & 1.000 \\
        Ensemble & 1.000 & 1.000 & 1.000 & 1.000 & 1.000 \\
        BBB      & 1.000 & 1.000 & 1.000 & 1.000 & 1.000 \\
        CP       & 0.988 & 0.999 & 1.000 & 1.000 & 1.000 \\
        SWAG     & \textbf{---} & \textbf{---} & 1.000 & 1.000 & 1.000 \\
        \midrule
        \multicolumn{6}{l}{\textit{Kin8nm}} \\
        MCD      & 1.000 & 1.000 & 1.000 & 1.000 & 1.000 \\
        Ensemble & 1.000 & 1.000 & 1.000 & 0.988 & 0.744 \\
        BBB      & 1.000 & 1.000 & 1.000 & 1.000 & 1.000 \\
        CP       & 1.000 & 1.000 & 1.000 & 1.000 & 1.000 \\
        SWAG     & \textbf{---} & \textbf{---} & \textbf{---} & 1.000 & 1.000 \\
        \midrule
        \multicolumn{6}{l}{\textit{Yacht Hydrodynamics}} \\
        MCD      & 1.000 & 1.000 & 0.954 & \textbf{---} & \textbf{---} \\
        Ensemble & 1.000 & 1.000 & 1.000 & \textbf{---} & \textbf{---} \\
        BBB      & 1.000 & 1.000 & 1.000 & \textbf{---} & \textbf{---} \\
        CP       & 0.000 & 0.000 & 0.000 & \textbf{---} & \textbf{---} \\
        \bottomrule
    \end{tabular}
\end{table}

\subsection{Real Dataset Complete Results}
\label{app:real}

Tables~\ref{tab:real_crps_app} and~\ref{tab:real_picp_app} report
mean CRPS and PICP across all real datasets and training sizes.

\begin{table}[h]
    \caption{Mean $\pm$ standard deviation of CRPS on real datasets.
    Lower is better.}
    \label{tab:real_crps_app}
    \centering
    \small
    \setlength{\tabcolsep}{4pt}
    \begin{tabular}{lrrrrr}
        \toprule
        Method & $n=30$ & $n=50$ & $n=100$ & $n=200$ & $n=500$ \\
        \midrule
        \multicolumn{6}{l}{\textit{Energy Efficiency}} \\
        MAP
          & $0.267 \pm 0.151$ & $0.219 \pm 0.065$
          & $0.167 \pm 0.042$ & $0.135 \pm 0.027$ & --- \\
        MCD
          & $0.159 \pm 0.041$ & $0.139 \pm 0.030$
          & $0.107 \pm 0.016$ & $0.095 \pm 0.010$ & --- \\
        Ensemble
          & $0.212 \pm 0.161$ & $0.164 \pm 0.040$
          & $0.113 \pm 0.017$ & $0.090 \pm 0.012$ & --- \\
        BBB
          & $0.198 \pm 0.060$ & $0.160 \pm 0.044$
          & $0.116 \pm 0.018$ & $0.091 \pm 0.008$ & --- \\
        CP
          & $0.368 \pm 0.149$ & $0.314 \pm 0.131$
          & $0.189 \pm 0.080$ & $0.127 \pm 0.028$ & --- \\
        SWAG
          & $1.261 \pm 1.843$ & $0.675 \pm 0.965$
          & $0.421 \pm 0.447$ & $0.351 \pm 0.372$ & --- \\
        \midrule
        \multicolumn{6}{l}{\textit{Concrete Compressive Strength}} \\
        MAP
          & $0.604 \pm 0.117$ & $0.561 \pm 0.108$
          & $0.522 \pm 0.101$ & $0.510 \pm 0.085$ & $0.448 \pm 0.045$ \\
        MCD
          & $0.454 \pm 0.079$ & $0.432 \pm 0.073$
          & $0.409 \pm 0.071$ & $0.423 \pm 0.043$ & $0.405 \pm 0.025$ \\
        Ensemble
          & $0.477 \pm 0.087$ & $0.451 \pm 0.112$
          & $0.423 \pm 0.077$ & $0.422 \pm 0.063$ & $0.416 \pm 0.037$ \\
        BBB
          & $0.465 \pm 0.079$ & $0.427 \pm 0.072$
          & $0.380 \pm 0.057$ & $0.388 \pm 0.046$ & $0.366 \pm 0.025$ \\
        CP
          & $0.552 \pm 0.104$ & $0.486 \pm 0.116$
          & $0.407 \pm 0.074$ & $0.394 \pm 0.067$ & $0.400 \pm 0.041$ \\
        SWAG
          & $0.772 \pm 0.742$ & $0.431 \pm 0.088$
          & $0.401 \pm 0.171$ & $0.397 \pm 0.067$ & $0.413 \pm 0.052$ \\
        \midrule
        \multicolumn{6}{l}{\textit{Kin8nm}} \\
        MAP
          & $0.742 \pm 0.061$ & $0.684 \pm 0.047$
          & $0.584 \pm 0.044$ & $0.527 \pm 0.123$ & $0.563 \pm 0.186$ \\
        MCD
          & $0.580 \pm 0.044$ & $0.511 \pm 0.028$
          & $0.427 \pm 0.025$ & $0.334 \pm 0.018$ & $0.251 \pm 0.007$ \\
        Ensemble
          & $0.608 \pm 0.056$ & $0.555 \pm 0.043$
          & $0.466 \pm 0.031$ & $0.469 \pm 0.110$ & $0.537 \pm 0.154$ \\
        BBB
          & $0.635 \pm 0.046$ & $0.571 \pm 0.036$
          & $0.484 \pm 0.030$ & $0.370 \pm 0.022$ & $0.259 \pm 0.010$ \\
        CP
          & $0.594 \pm 0.061$ & $0.552 \pm 0.041$
          & $0.479 \pm 0.029$ & $0.410 \pm 0.020$ & $0.319 \pm 0.018$ \\
        SWAG
          & $1.551 \pm 1.448$ & $1.029 \pm 1.047$
          & $0.583 \pm 0.212$ & $0.374 \pm 0.027$ & $0.277 \pm 0.009$ \\
        \midrule
        \multicolumn{6}{l}{\textit{Yacht Hydrodynamics}} \\
        MAP
          & $0.326 \pm 0.075$ & $0.258 \pm 0.051$
          & $0.197 \pm 0.032$ & --- & --- \\
        MCD
          & $0.267 \pm 0.060$ & $0.220 \pm 0.050$
          & $0.179 \pm 0.066$ & --- & --- \\
        Ensemble
          & $0.254 \pm 0.055$ & $0.194 \pm 0.054$
          & $0.122 \pm 0.031$ & --- & --- \\
        BBB
          & $0.260 \pm 0.058$ & $0.189 \pm 0.041$
          & $0.140 \pm 0.020$ & --- & --- \\
        CP
          & $0.449 \pm 0.089$ & $0.363 \pm 0.075$
          & $0.275 \pm 0.060$ & --- & --- \\
        SWAG
          & $1.699 \pm 2.468$ & $0.954 \pm 1.302$
          & $0.804 \pm 1.015$ & --- & --- \\
        \bottomrule
    \end{tabular}
\end{table}

\begin{table}[h]
    \caption{Mean PICP on real datasets. Target value is 0.90.}
    \label{tab:real_picp_app}
    \centering
    \small
    \setlength{\tabcolsep}{4pt}
    \begin{tabular}{lrrrrr}
        \toprule
        Method & $n=30$ & $n=50$ & $n=100$ & $n=200$ & $n=500$ \\
        \midrule
        \multicolumn{6}{l}{\textit{Energy Efficiency}} \\
        MAP      & 0.001 & 0.001 & 0.006 & 0.020 & --- \\
        MCD      & 0.762 & 0.803 & 0.880 & 0.918 & --- \\
        Ensemble & 0.595 & 0.664 & 0.770 & 0.813 & --- \\
        BBB      & 0.516 & 0.605 & 0.743 & 0.859 & --- \\
        CP       & 0.939 & 0.954 & 0.941 & 0.922 & --- \\
        SWAG     & 0.912 & 0.911 & 0.934 & 0.922 & --- \\
        \midrule
        \multicolumn{6}{l}{\textit{Concrete Compressive Strength}} \\
        MAP      & 0.000 & 0.001 & 0.013 & 0.046 & 0.108 \\
        MCD      & 0.468 & 0.515 & 0.571 & 0.567 & 0.586 \\
        Ensemble & 0.471 & 0.486 & 0.463 & 0.432 & 0.389 \\
        BBB      & 0.357 & 0.398 & 0.471 & 0.499 & 0.554 \\
        CP       & 0.932 & 0.934 & 0.858 & 0.747 & 0.592 \\
        SWAG     & 0.904 & 0.840 & 0.726 & 0.567 & 0.508 \\
        \midrule
        \multicolumn{6}{l}{\textit{Kin8nm}} \\
        MAP      & 0.000 & 0.000 & 0.001 & 0.023 & 0.202 \\
        MCD      & 0.334 & 0.406 & 0.524 & 0.686 & 0.854 \\
        Ensemble & 0.392 & 0.416 & 0.476 & 0.568 & 0.691 \\
        BBB      & 0.213 & 0.249 & 0.323 & 0.482 & 0.721 \\
        CP       & 0.936 & 0.927 & 0.907 & 0.904 & 0.900 \\
        SWAG     & 0.861 & 0.873 & 0.853 & 0.709 & 0.743 \\
        \midrule
        \multicolumn{6}{l}{\textit{Yacht Hydrodynamics}} \\
        MAP      & 0.001 & 0.000 & 0.001 & --- & --- \\
        MCD      & 0.623 & 0.632 & 0.757 & --- & --- \\
        Ensemble & 0.756 & 0.833 & 0.890 & --- & --- \\
        BBB      & 0.589 & 0.724 & 0.803 & --- & --- \\
        CP       & 0.920 & 0.910 & 0.707 & --- & --- \\
        SWAG     & 0.864 & 0.913 & 0.942 & --- & --- \\
        \bottomrule
    \end{tabular}
\end{table}

\end{document}